%% file: paper.tex
\documentclass[conference]{IEEEtran}
\IEEEoverridecommandlockouts

\usepackage{xspace}
\usepackage{amsmath}
\usepackage{amsmath,amssymb,amsfonts,amsfonts,amsthm}
\usepackage{algorithm}
\usepackage[noend]{algorithmic}
\usepackage[caption=false,font=footnotesize]{subfig}	
\usepackage{framed} 
\usepackage[]{epsfig,color}
\usepackage[nice]{nicefrac}
\usepackage{multirow}
\usepackage{url}

\input{macros.tex}

\begin{document}
%

\title{Sparsification of Motion-Planning Roadmaps by Edge Contraction}


%

\author
{
	\IEEEauthorblockN{Doron Shaharabani\IEEEauthorrefmark{1},
										Oren Salzman\IEEEauthorrefmark{1},  
										Pankaj K. Agarwal\IEEEauthorrefmark{2} and
										Dan Halperin\IEEEauthorrefmark{1}}
	\IEEEauthorblockA{
		\IEEEauthorrefmark{1}
			Balvatnic School of Computer Science,
			Tel-Aviv University, Israel}
		\thanks{Work by D.S., O.S., and D.H. has been supported in part by the 7th 
		Framework Programme for Research of the European Commission, under
    FET-Open grant number 255827 (CGL---Computational Geometry
    Learning), by the Israel Science Foundation (grant no.
    1102/11), and by the Hermann Minkowski--Minerva Center for
    Geometry at Tel Aviv University.}
  \IEEEauthorblockA{
		\IEEEauthorrefmark{2}	
			Department of Computer Science, 
			Duke University, USA}
		\thanks{Work by P.A. was supported by NSF under grants IIS-07-13498, 
						CCF-09-40671, CCF-10-12254, and CCF-11-61359,by ARO grants
						W911NF-07-1-0376 and W911NF-08-1-0452, and by an ARL award
						W9132V-11-C-0003.}
}


\maketitle

\input {abstract.tex}

%
\IEEEpeerreviewmaketitle

\input{related_work.tex}
\input{algorithmic_framework.tex}
\input{algorithmic_details.tex}
\input{experimntal_results.tex}
\input{discussion.tex}


\bibliographystyle{IEEEtran}
\bibliography{IEEEabrv,bibliography}

\arxiv
{
}
{
\appendix
\input{additional_experimntal_results.tex}
}

\end{document}

%% file: macros.tex
\newcommand{\calC}{\ensuremath{\mathcal{C}}\xspace}

\newcommand{\calG}{\ensuremath{\mathcal{G}}\xspace}

\newcommand{\Cfree}{\ensuremath{\calC_{\rm free}}\xspace}

\newcommand{\Cs}{C-space\xspace}
\newcommand{\Css}{C-spaces\xspace}

\newcommand{\ignore}[1]{}

\newcommand{\arxiv}[2]{#2}

%% file: abstract.tex

\begin{abstract}
We present Roadmap Sparsification by Edge Contraction (RSEC), a simple and effective algorithm for reducing the size of a motion-planning roadmap.
The algorithm exhibits minimal effect on the quality of paths that can be extracted from the new roadmap.
The primitive operation used by RSEC is \emph{edge contraction}---the contraction of a roadmap edge to a single vertex and the connection of the new vertex to the neighboring vertices of the contracted edge.
For certain scenarios, we compress more than 98\% of the edges and vertices at the cost of degradation of average shortest path length by at most 2\%.
\end{abstract}

%% file: related_work.tex
\section{Introduction}
\label{sec:related_work}

The introduction of sampling-based planners~\cite{HLM99, KSLO96, KL00} enabled solving motion-planning problems that were previously infeasible~\cite{CBHKKLT05}. 
Specifically, for multi-query scenarios, planners such as the Probabilistic  Roadmap Planner~(PRM)~\cite{KSLO96} approximate the connectivity of the free space~(\Cfree) by taking random samples from the Configuration Space~(\Cs) and connecting near-by free configurations when possible.
The resulting data structure, the \emph{roadmap}, is a graph where vertices represent configurations in \Cfree and edges are collision-free paths connecting two such configurations.

Answering a motion-planning query using such a roadmap is subdivided into 
(i)~connecting the source and target configurations of the query to the roadmap, and 
(ii)~finding a path in the roadmap between the connection points. 
Thus, a roadmap should cover the \Cs (\emph{coverage} property) and be connected when the \Cs is connected (\emph{connectivity} property).
Typically, a path of \emph{high quality} is desired where quality can be measured in terms of length, clearance, smoothness, energy, to mention a few criteria, or some combination of the above. As the connectivity property alone does not guarantee high-quality paths~\cite{NRH10,KF11} additional work is required.

Smoothing is a common practice that may improve the quality of a single path. In particular when applied to a path extracted from the output roadmap of an algorithm that produces paths deformable to optimal ones 
(see, e.g.,~\cite{SBDDH02, JS06}). However smoothing is costly and needs to be applied for every query.
Nieuwenhuisen and Overmars~\cite{NO04} suggested adding useful cycles to the roadmap in order to improve the resulting quality of the paths. 
Raveh et al.~\cite{REH11} proposed combining the output of several different planners. 
Recent work by Karaman and Frazzoli~\cite{KF11} introduced several sampling based algorithms, including PRM*, such that asymptotically, the solution returned by the roadmap almost surely converges to an optimum path (with regards to some quality measure). 
This desired behavior comes at the expense of increasing the number of neighbors each node is connected to, as the number of samples increases.

Thus, motion-planning algorithms that create high-quality roadmaps result in large, dense graphs. This may be undesirable due to prohibitive storage requirements and long online query processing time. 
We seek a compact representation of the roadmap graph without sacrificing the desirable guarantees on path quality.

\subsection{Related work}
Geraerts and Overmars~\cite{GO07} suggested an algorithm for creating small roadmaps of high-quality but their technique is limited to two- and 
three-dimensional \Css. The work on graph spanners~\cite{PS89,NS07} is closely related to our problem of computing a compact representation of a roadmap graph.
Formally, a $t$-spanner of a graph $\calG =(V,E)$ is a sparse subgraph 
$\calG' =(V,E'\subseteq E)$, where the shortest path between any pair of points in $\calG'$ is no longer than $t$ times the shortest path between them in \calG. The parameter $t$ is referred to as the \emph{stretch} of 
the spanner. It is well known that small size $(1+\epsilon)$-spanners exist for complete Euclidean graphs~\cite{NS07,Aryaetal,ES12}. Spanners have been successfully used to compute collision-free approximate shortest paths amid 
obstacles in 2D and 3D or to compute them on a surface~\cite{Cl87,AMS05,HP99,VA00}. In the context of roadmap constructions,
Marble and Bekris~\cite{MB11-IROS} introduced the notion of 
\emph{Asymptotically near-optimal roadmaps}---roadmaps that are guaranteed to return a path whose quality is within a guaranteed factor of the optimal 
path. They apply a graph-spanner algorithm  to an existing roadmap to 
reduce its size.  Recently, graph-spanner algorithms have also been 
incorporated in the construction phase of the roadmap itself~\cite{DKB12, MB11-ISRR, MB12}. 

A drawback of the spanner based approach is that it only reduces the number of edges of the graph and does not remove any of its vertices. In the context of roadmaps, it will be useful to remove redundant vertices as well and to construct a small-size graph in the free space. Such an approach was recently 
proposed for computing approximate shortest paths for a point robot amid convex 
obstacles in two or three dimensions~\cite{ASY09}, but it is not clear how to extend this approach to higher dimensional configuration spaces.

The problem of computing a compact representation of a roadmap graph 
falls under the broad area of computing data summaries or computing a 
hierarchical representation of data. There has been extensive work in this area in the last decade because of the need to cope with big data sets.  
The work in computer graphics on computing a hierarchical representation 
of a surface, represented as a triangulated mesh, is perhaps the most 
closely related work to our approach. The surface-simplification problems asks 
for  simplifying the mesh---reducing its  size---while ensuring that the 
resulting surface approximates the original one within a prescribed error 
tolerance~\cite{LRbook}. Some versions of the optimal surface-simplification problem, the problem of computing a smallest-size surface, are known to be 
NP-Hard~\cite{AS98}, and 
several approximation algorithms and heuristics have been proposed. The 
practical algorithms progressively simplify the surface by modifying
its topology locally at each step, e.g., removing a vertex and retriangulating the surface, or contracting an edge. The \emph{edge-contraction} method has by now become the most popular method for simplifying a surface~\cite{HDDMS93,GH97,Edelsbrunner01}.

\subsection{Contribution}
In this work, we adapt the widely used \emph{edge-contraction} technique for surface simplification to roadmap simplification. 
We suggest a simple algorithm to sparsify a roadmap---Roadmap Sparsification by Edge Contraction (RSEC) where edge contraction is the primitive operation.
Our algorithm often exhibits little degradation in path quality when compared to the original graph while providing very sparse graphs. 
In contrast to the algorithm of Marble and Bekris~\cite{MB11-IROS} in which the set of vertices remains intact, RSEC dramatically reduces the number of vertices.

In Section~\ref{sec:algorithm} we present the algorithmic framework 
and in Section~\ref{sec:algorithmic_details} we cover the implementation details.
Section~\ref{sec:experimental results} presents experimental results comparing
our implementation with several alternatives.
For certain scenarios, we compress more than 98\% of the edges and vertices while causing degradation of average path length by at most 2\%.
Additionally, we compare RSEC with the algorithm presented by Marble and Bekris~\cite{MB11-IROS}. We show that our algorithm produces paths of higher quality than theirs when applying the same compression rate while also being able to surpass the maximum compression rate achieved by their algorithm.
We conclude with a discussion and suggestions for further research in Section~\ref{sec:discussion}.

%% file: algorithmic_framework.tex
\section{Algorithmic framework}
\label{sec:algorithm}

Let $\calG = (V, E)$ be an undirected input graph, which is the output of a PRM-type algorithm approximating \Cfree. Informally, we wish to construct a graph $\calG' = (V', E')$ such that $\calG'$ is a more compact approximation of \Cfree while maintaining 
the coverage of the roadmap and 
the quality of the approximation provided by $\calG$. 
In this work we concentrate on the quality measure of path length.

We denote the set of all neighbors of a vertex $v \in V$ by:
\[
\text{neighbors}(v) = \{ u \in V | (u,v) \in E \}.
\]
Given an edge $(u,v)$ and a point $p \in \Cfree$, we define an \emph{edge contraction}  of $(u,v)$ to $p$ as the following process: 
Adding $p$ as a new vertex, adding an edge $(w,p)$ for each vertex $w \in \{ \text{neighbors}(u) \cup \text{neighbors}(v) \}$ and
removing the vertices $u,v$ and all their neighboring edges from the graph. 
We say that $u$ and $v$ were contracted to $p$ and define:
\[
\text{parent}(v')  = \{v| v \text{ was contracted to } v' \}.
\]
Using this definition, we can recursively define that a vertex $v$ is an \emph{ancestor} of a vertex $v'$ if either: (i)~$v$ is a parent of $v'$ or (ii)~there exists a vertex $u$ such that $u$ is the parent of $v'$ and $v$ is an ancestor of $u$. 

%
An edge contraction is considered \emph{legal} if each new edge is collision-free.
This is detailed in Algorithm~\ref{alg:is_contractible}, which performs the validity check that an edge $(u,v)$ can be contracted to a point $p$ using a local planner\footnote{A local planner is a predicate that determines if there exists a collision-free path between two configurations.} and in Algorithm~\ref{alg:contraction}, which updates the graph \calG after the contraction. 

Connecting an edge $(u,v)$ to a graph~\calG reduces the size of~\calG---the number of vertices decreases by one and the number of edges decreases by at least one (the edge $(u,v)$). Additionally, for every common neighbor $w$ of $u$ and $v$, the edges $(u,w)$ and $(v,w)$ merge into a single edge $(p,w)$.

\begin{algorithm}[t,b]
\caption{is\_contractible ($\calG(E,V), (u,v), p$)}
\label{alg:is_contractible}
\begin{algorithmic}[1]
 \FORALL {$w \in \{ \calG.\text{neighbors}(v) \cup 
 										\calG.\text{neighbors}(u) \}$}
 		\IF {local\_planner($w,p$) = FAILURE} 		
 			\RETURN FAILURE
 		\ENDIF
 	\ENDFOR
 	
	\RETURN SUCCESS
	
\end{algorithmic}
\end{algorithm}

\begin{algorithm}[t,b]
\caption{edge\_contraction ($\calG(E,V), (u,v), p$)}
\label{alg:contraction}
\begin{algorithmic}[1]
 \IF {is\_contractible($\calG,(u,v), p$)}
	 \FORALL {$v' \in \{ \calG.\text{neighbors}(v)\}$} 
	 		\STATE $E \leftarrow E \cup\ \{(v',p)\}$
	 		\STATE $E \leftarrow E  \setminus\ \{(v',v)\}$
	 	\ENDFOR
	 	\FORALL {$u' \in \{ \calG.\text{neighbors}(u)\}$} 
	 		\STATE $E \leftarrow E \cup\ \{(u',p)\}$	 	
	 		\STATE $E \leftarrow E \setminus\ \{(u',u)\}$	 	
	 	\ENDFOR
	 	\STATE $V \leftarrow V \cup \{p\}$
	 	\STATE $V \leftarrow V \setminus \{v,u\}$
	 		
		\RETURN SUCCESS
 \ELSE
 	\RETURN FAILURE
 \ENDIF
 	
\end{algorithmic}
\end{algorithm}

Our algorithm, Roadmap Sparsification by Edge Contraction (RSEC), performs a series of legal edge contractions as detailed in Algorithm~\ref{alg:sparsification}. The algorithm maintains an order on the edges considered for contraction by using a priority queue ordered according to some weight function (line~$2$). At each step the current edge is popped out of the queue (line~$4$) and a contraction point is computed (line~$5$). If the prospective contraction point is valid and the edge contraction was successful (lines~$6,7$), the queue is updated, as the weights of edges may have changed~(line~$8$). The process terminates when the queue is empty.

\begin{algorithm}[t,b]
\caption{RSEC ($\calG(E,V)$)}
\label{alg:sparsification}
\begin{algorithmic}[1]
 \STATE $\calG' \leftarrow \calG$
 \STATE $Q \leftarrow \text{initialize\_queue}(E)$
 	
 \WHILE {not\_empty($Q$)}
 	\STATE $e \leftarrow Q.\text{pop\_head}$
 	\STATE $p \leftarrow \text{get\_contraction\_point(e)}$
 	
 	\IF {$p \neq $NIL 	}
 		\IF {edge\_contraction ($\calG', e, p$) == SUCCESS}
 			\STATE $Q \leftarrow \text{update\_queue}(E')$
 		\ENDIF
 	\ENDIF
 	
 \ENDWHILE
 	
\end{algorithmic}
\end{algorithm}

\begin{algorithm}[t,b]
\caption{get\_contraction\_point ($e=(u,v)$)}
\label{alg:contraction_point}
\begin{algorithmic}[1]
 \STATE $p \leftarrow \text{random\_point}(u,v)$
 \IF {$ \text{collision\_detector} (p) \neq$ FREE}
 	\RETURN NIL
 \ENDIF
  
 \FORALL {$w \in u.\text{ancestors}() \ \bigcup \ v.\text{ancestors}() $}
 	\IF {$\text{distance} (w, p) > d$}
 		\RETURN NIL
 	\ENDIF
 \ENDFOR
 
 \STATE $p.\text{ancestors}() \leftarrow 
 						u.\text{ancestors}() \ \bigcup \ v.\text{ancestors}()$
 \RETURN $p$
 
\end{algorithmic}
\end{algorithm}

%% file: algorithmic_details.tex
\section{Algorithmic details}
\label{sec:algorithmic_details}
Algorithm~\ref{alg:sparsification} presents RSEC as a general, modular framework. As such, some technical details still need to be addressed in order to complete its description. Specifically, what point should an edge be contracted to? How do we order the edges in our queue to obtain a sparse graph of high quality? 

\subsection{Contraction point choice}
When contracting an edge $e$ to a point $p$, several alternatives regarding the choice of $p$ come to mind:
(i)~The midpoint of~$e$, 
(ii)~an incident vertex of~$e$ or 
(iii)~a random point along~$e$. 
One may also consider several contraction points when testing whether an edge is contractible. 

The advantage of alternative~(ii) is that the neighboring edges of the vertex chosen to be the contraction point are not altered. This reduces dramatically the number of collision checks needed by the algorithm. 
Alternatives~(i) and~(iii), on the other hand, seem to distribute naturally the location of the contraction point as the average of its ancestors.
We chose the contraction point to be alternative~(iii), random point along the edge, as we observed higher path degradation in comparison to the other alternatives.

\subsection{Quality-driven constraint}
As we wish to preserve the quality and connectivity of the output graph we add the constraint that vertices in the new graph are not too far from their ancestors.
Then, given a \emph{drift bound} $d>0$ and a distance function $dist$, we maintain the following invariant in our algorithm: 
\begin{framed}
\textbf{Bounded drift invariant} - 
For every vertex $v' \in V'$ 
and for every $v \in \text{ancestor}(v')$,
$dist(v', v) \leq d$. 
\end{framed}

Line~5 of Algorithm~\ref{alg:sparsification} performs a call to a subroutine get\_contraction\_point($e$). The subroutine verifies that the contraction point does not violate the bounded drift invariant. Algorithm~\ref{alg:contraction_point} demonstrates an implementation of this subroutine.

We note that in practice we normalize the drift bound to be $\frac{d}{a}$ where $a$ is the length of the workspace bounding box diagonal. 
In the rest of the paper we use the term drift bound to refer to the normalized drift bound.

%
%
%

\subsection{Edge ordering}
We suggest to choose the edges next edge~$e=(u,v)$ for contraction as follows:
We order (from low to high) the edges according to the sum of degrees of the vertices incident to the edge, 
namely degree$(u)$ $+$ degree$(v)$, 
and name the heuristic  \emph{deg\_sum}.
The motivation for this heuristics comes from the fact that vertices with low degree impose less validity checks (and hence less constraints) when checking whether an edge is contactable. This not only reduces the computation time, but more importantly, increases the probability of a valid contraction.

In Section~\ref{sec:experimental results} we evaluate our heuristic by suggesting several alternatives and comparing them to deg\_sum.  
Indeed, in the experimental results deg\_sum outperforms the alternative heuristics.

\begin{figure*}[t,b]
  \centering
  \subfloat
   [\sf ]
   {
			\includegraphics[width=0.2\textwidth]{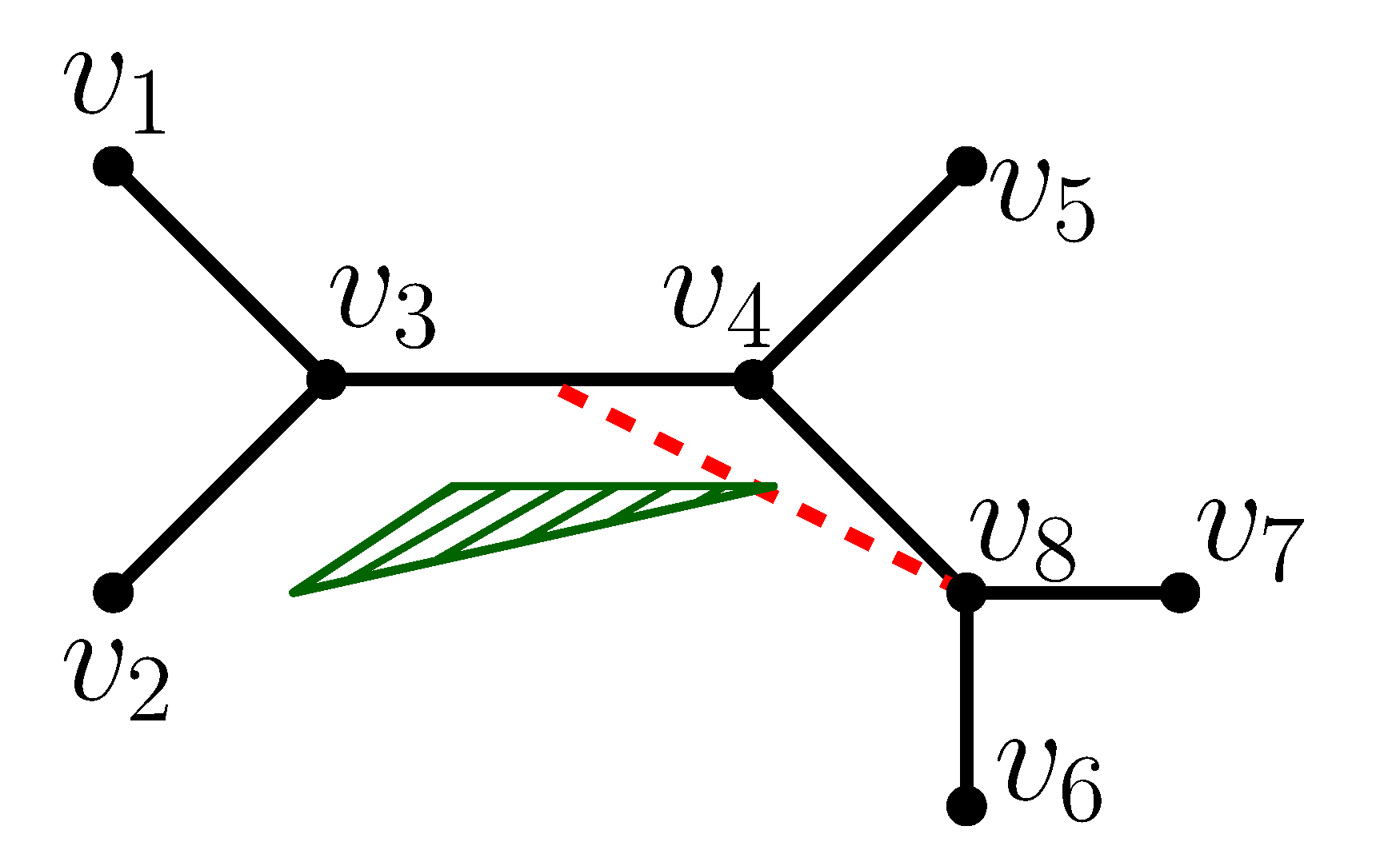}
   }
   \hspace{3mm}
  \subfloat
   [\sf ]
   {
    \includegraphics[width=0.2\textwidth]{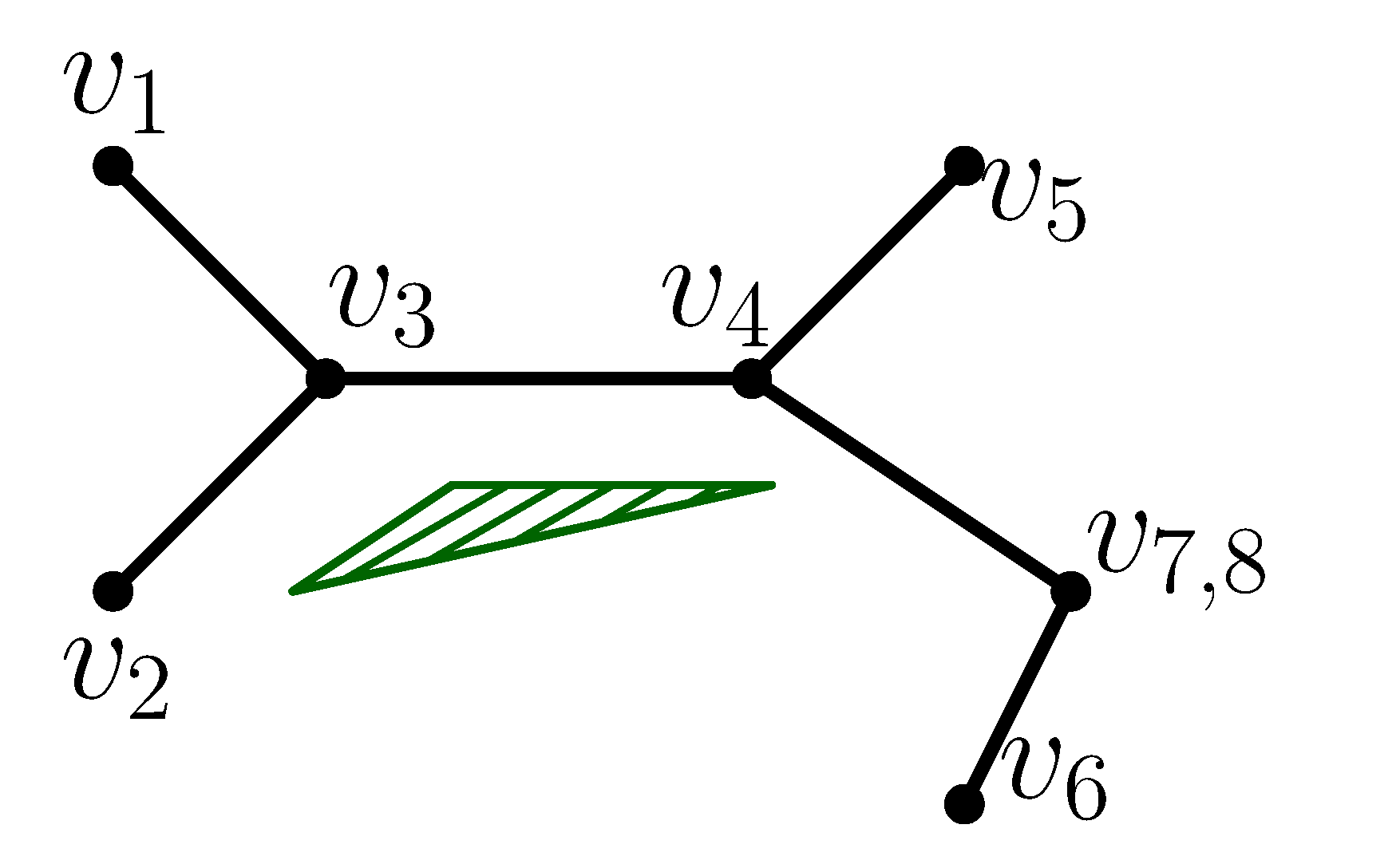}
   }
   \hspace{3mm}
  \subfloat
   [\sf ]
   {
    \includegraphics[width=0.2\textwidth]{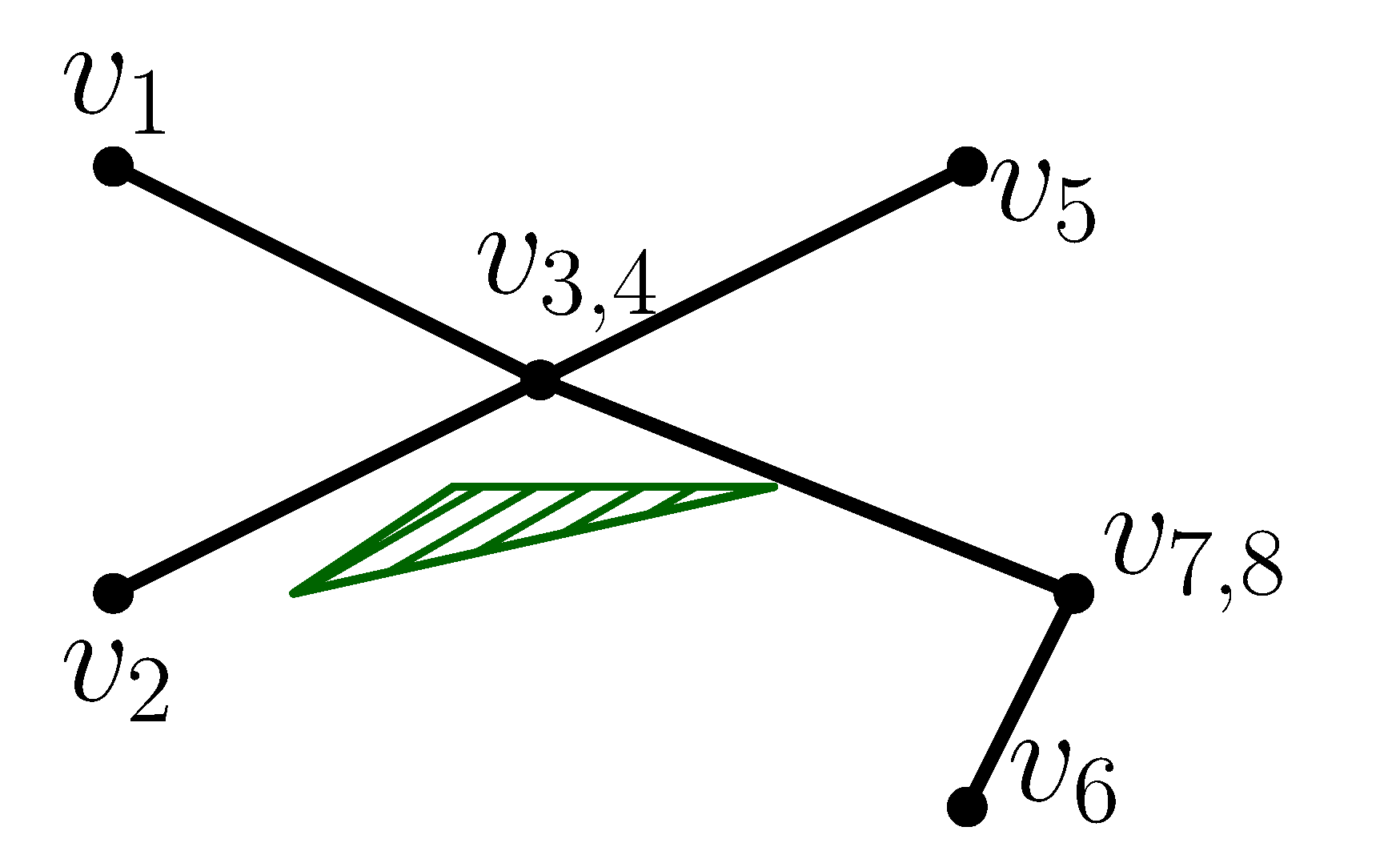}
   }
	\caption{\sf 	Scenario where re-inserting an edge to the priority queue 
								enables additional edge contractions.
								(a) Edge $(v_3, v_4)$ is considered for contraction but the 
  							contraction is not valid and the edge is removed from the 
  							queue as the prospective edge from $v_8$ will collide with the 
  							green obstacle (demonstrated by the dashed red line).
  							(b) The edge $(v_7, v_8)$ is contracted to the point $v_{7,8}$ 
  							(c) Re-inserting the edge $(v_3, v_4)$ to the queue allows it 
  							to be contracted to the point~$v_{3,4}$.}
  \label{fig:reinsertion}
\end{figure*}

An additional implementation issue that should be addressed regards the question ``should edges be re-inserted into the priority queue?''
An edge $(u,v)$ that was removed from the queue because it could not be contracted could possibly be contracted at a later stage of the algorithm. Such an event could occur when a vertex that is adjacent to either $u$ or $v$ moves as a result of a contraction operation on a different edge. An example of such a scenario is depicted in Figure~\ref{fig:reinsertion}.
Obviously, re-inserting these edges is costly with regards to runtime but improves the compression achieved. 
As we did not consider processing time to be a major constraint, we allowed edges to be re-inserted to the queue.

%% file: experimntal_results.tex
	\section{Evaluation}
\label{sec:experimental results}
\begin{figure*}[t,b]
  \centering
  \subfloat
   [\sf Easy]
   {
    \includegraphics[width=0.2\textwidth]{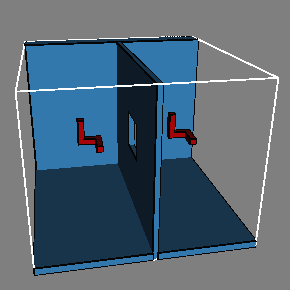}
   }
   \hspace{1mm}
  \subfloat
   [\sf Cubicles]
   {
    \includegraphics[width=0.2\textwidth]{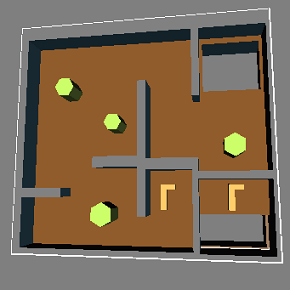}
   }
   \hspace{1mm}
  \subfloat
   [\sf Bug trap]
   {
    \includegraphics[width=0.2\textwidth]{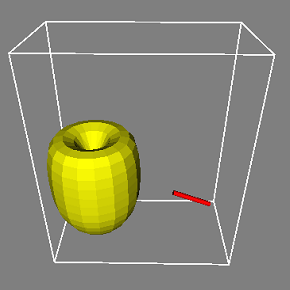}
   }
   \hspace{1mm}
  \subfloat
   [\sf Alpha puzzle]
   {
    \includegraphics[width=0.2\textwidth]{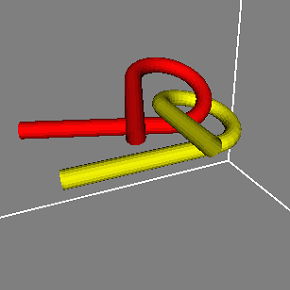}
   }
  \caption{\sf 	Environments used for the experiments. 
  							The robot is depicted in several configurations in each 
  							environment. 
  							All scenarios were taken from the OMPL~\cite{SMK12} 
  							distribution.}
  \vspace {-5mm}
  \label{fig:scenarios}
\end{figure*}

We evaluated our algorithm by running the PRM*~\cite{KF11} algorithm on several scenarios to obtain initial roadmaps. 
We first compared our heuristic for ordering the edges with several alternatives. Then we compared RSEC with the spanner-based sparcification algorithm by Marble and Bekris~\cite{MB11-IROS} (which we will call SPANNER).
We chose to compare our algorithm to SPANNER as it is, to the best of our knowledge, the only offline algorithm for reducing roadmaps.

We measure the quality of a roadmap in terms of path length and report two values: 
(i)~\emph{path degradation} and (ii)~\emph{compression factor}.
The former is the ratio of the average shortest path length in the sparse graph and the average shortest path length in the original graph for random queries.
The latter is the ratio $\frac{|\calG|}{|\calG'|}$ where $|\calG|$ be the size\footnote{We assume that each vertex stores the coordinates of the configuration it represents (which is proportional in size to the number of degrees of freedom) and each edge stores the indices of the vertices it connects and its weight. 
Thus for our scenarios in SE03 we measure the size of a roadmap~$\calG(V,E)$ as $|\calG| =  6|V| + 3|E|$.} 
of the original roadmap and $|\calG'|$ be the size of the compressed roadmap.

All experiments were run using the Open Motion Planning Library (OMPL)~\cite{SMK12} on a 3.4GHz Intel Core i7 processor with 8GB of memory. 
We used several scenarios provided by the OMPL distribution. The scenarios are depicted in Figure~\ref{fig:scenarios}.

\subsection{Heuristic evaluation}
\label{subsec:eval}
In Section~\ref{sec:algorithmic_details} we described our edge ordering heuristic, namely deg\_sum.
Several natural alternatives may come to mind:
The first, termed \emph{compressibility}, chooses the next edge~$e=(u,v)$ to contract according to how many edges could potentially be eliminated from the graph by contracting $e$ into a single point. 
The number of edges that will be eliminated (except from the contracted edge) is the number of common neighbors of $u$ and $v$. 
We normalize this value by dividing by the total number of neighbors of $u$ and $v$ and define the compreassbility of $(u,v)$ as:
\[
	\text{compressibility}(u,v)= 
		\frac	{| \text{neighbors}(u) \cap \text{neighbors}(v) |}
					{| \text{neighbors}(u) \cup \text{neighbors}(v) |}
	.
\]

The second heuristic that one may think of chooses the next edge to contract according to its \emph{clearance}, namely the distance of the edge from the closest obstacle (ordered from high clearance to low). This is motivated by the fact that edges of high-clearance may be good candidates for successful contractions.
A third option may be a simple first in first out (FIFO) ordering scheme.

\begin{figure}[h]
  \centering
 	\includegraphics[width=0.33\textwidth]
 		{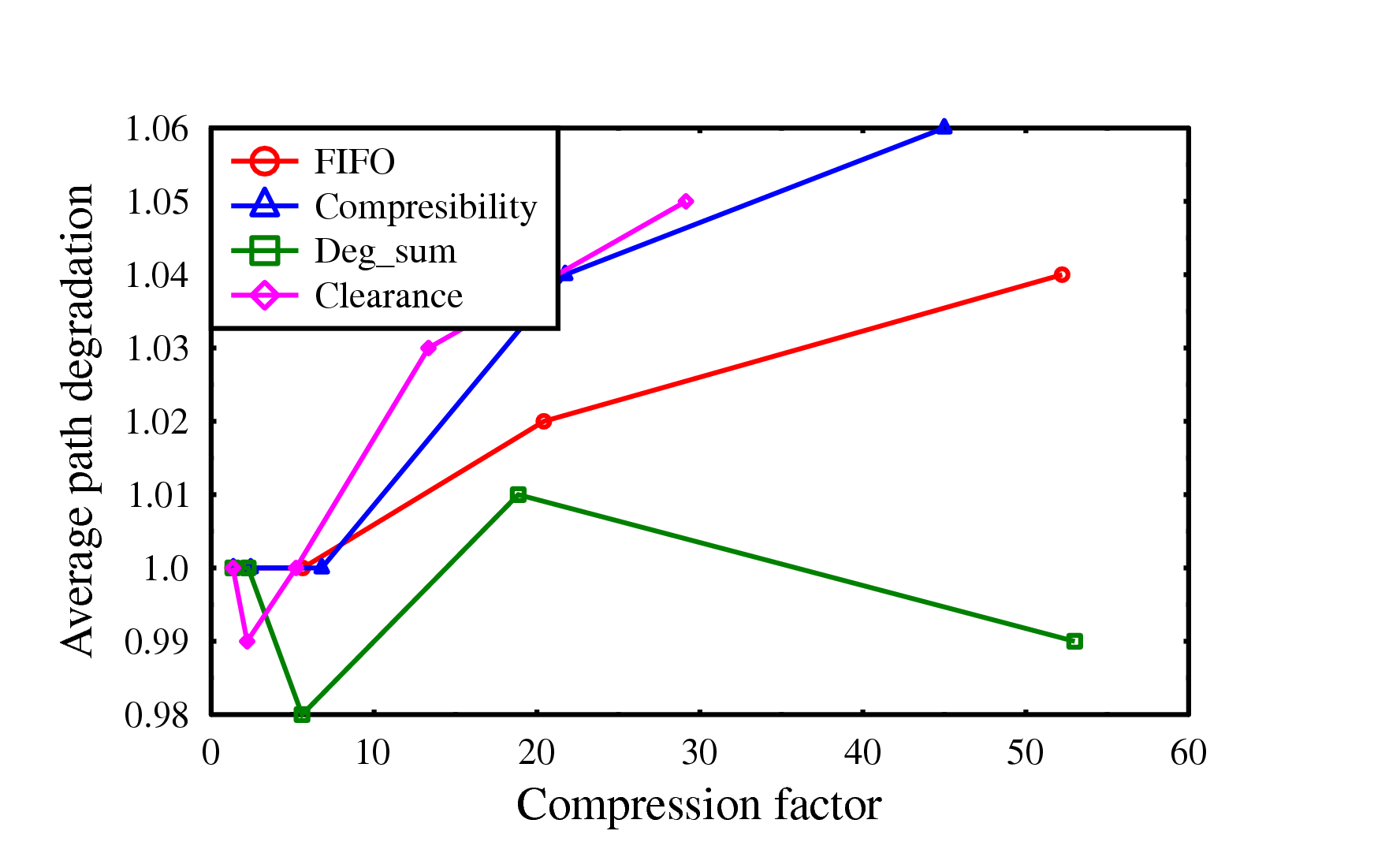}
   \caption{\sf Comparison of different heuristics on the cubicles scenario.}
  \label{fig:heuristics_comparison}
\end{figure}

We ran the alternative implementations on roadmaps containing 5,000 vertices and approximately 50,000 edges. The results for the Cubicles scenario are summarized in
Figure~\ref{fig:heuristics_comparison}.

For low compression factors, all heuristics exhibit no degradation with regards to path quality. A possible explanation is that for small perturbations, our algorithm may be seen as a smoothing of the graph. 
For high compression factors, the path degradation is very low for all heuristics, where deg\_sum seems to be the best choice. 

One can see that deg\_sum and FIFO achieve the highest compression factors. 
In order to explain this, we note that vertices with high degrees are more likely to fail a contraction operation, thus providing a negative influence to the overall compression achieved by RSEC. 
The clearance heuristic prioritizes edges with high clearance, resulting in contraction operations being applied repeatedly to high-clearance areas of the roadmap.
As each contraction operation creates a vertex with a higher degree than its ancestors', biasing the contraction operations to specific areas of the graph at a time is likely to create high degree vertices. 
In a similar manner, the compressibility is likely to focus the contractions on highly connected areas. 
On the other hand, the FIFO heuristic performs the contraction operations without prioritizing any specific areas of the graph, and the deg\_sum heuristic aims to prevent high-degree vertices by concentrating on edges who's incident vertices have a low degree

Table~\ref{tbl:heuristic_comprisons} summarizes the algorithm's results on all scenarios for the deg\_sum heuristic.
\arxiv
{
For the complete set of results, omitted due to lack of space, we refer the reader to~\cite{arxiv}.
}
{
For the complete set of results, we refer the reader to the appendix.
}

\begin{table*}[t,b]
\footnotesize
\begin{center}
	\begin{tabular}{|c c||c c c c c c||}
		\hline
		&	& &\multicolumn{5}{c||}{drift bound}\\
		Scenario 			
		&	&	original &	$d=0.1$ 	&	$d=0.2$	& $d=0.4$	& $d=0.8$ & $d=0.16$\\
		\hline
		{\multirow{4}{*}{Easy}} 
		&	average path length&  314.8&
								99.8\%	&	99.2\%		&	98.1\%	&	97.5\%	&	99.1\%	\\
		&	compression factor&		1&
								1.16		&	1.82	 		&	3.37		&	11.4 		&	17.6		\\
		&	vertices					&		5016&
								87.2\%	&	59.0\%		&	26.2\%	&	10.1\% 	&	3.6\%		\\
		&	edges							&		48866&
								85.7\%	&	54.0\%		&	30.4\%	&	14.0\%	&	6.1\%		\\
		\hline
		{\multirow{4}{*}{Cubicles}} 
		&	average path length&  472.4&
								99.9\%	&	99.5\%	&	98.1\%	&	101.2\%	&	99.2\%	\\
		&	compression factor&	  1&
								1.33		&	2.27 		&	5.58 		&	18.85		&	53		\\
		&	vertices					&	  5004&
								76.8\%	&	45.6\%	&	16.4\%	&	5.3\% 	&	2.0\%		\\
		&	edges							&	  45217&
								74.8\%	&	43.7\%	&	18.3\%	&	5.3\%		&	1.9\%		\\
		\hline
		{\multirow{4}{*}{Bug trap}} 
		&	average path length&	  41.8&
								100\%		&	99.6\%	&	97.6\%	&	95.1\%	&	93.8\%	\\
		&	compression factor&	  1&
								1.03		&	1.6	 		&	2.9 		&	7.2 		&	25.8		\\
		&	vertices					&	  5011&
								95.8\%	&	66.9\%	&	33.3\%	&	12.3\% 	&	3.4\%		\\
		&	edges							&  66053&
								97.1\%	&	61.6\%	&	33.4\%	&	14.1\%	&	3.9\%		\\		\hline
		{\multirow{4}{*}{Alpha puzzle}} 
		&	average path length&  127.8&
								99.9\%	&	99.3\%	&	98.1\%	&	97.1\%	&	97.6\%  \\
		&	compression factor&  1&
								1.09		&	1.74 		&	3.23 		&	6.57 		&	16.3		\\
		&	vertices					&  5053&
								92.1\%	&	61.3\%	&	28.3\%	&	11.6\%	&	4.1\%		\\
		&	edges							&	  62294&
								91.8\%	&	56.8\%	&	31.4\%	&	15.8\%	&	6.5\%		\\
		\hline		
	\end{tabular}		
\end{center}
\caption{\sf  Average path degradation and compression factor of RSEC 	
							with respect to the original graph using the deg\_sum 
							heuristic.
							The results are averaged over 5 runs and 150 random queries.}
\label{tbl:heuristic_comprisons}
\end{table*}

\subsection{Comparison with SPANNER}
When comparing RSEC with SPANNER, we used the same scenarios depicted in Figure~\ref{fig:scenarios} with two initial roadmap sizes. The first, which we call \emph{normal setting}, contains a roadmap with 5,000 vertices and approximately 50,000 edges. The second, which we call \emph{dense setting}, contains 20,000 vertices and approximately 1.2 million edges. The dense setting resembles the benchmark used by Marble and Bekris~\cite{MB11-IROS}. 
The crucial difference between our two settings is not the size but the average degree of each vertex (20 and 120 for the normal setting and dense setting, respectively).

\vspace{3mm}
\textbf{Roadmap Connectivity}
We desire that a sparsification algorithm retains some quality measures while not sacrificing the ability to connect queries to the roadmap.
Hence, we sampled 1000 random free configurations (playing the role of start or goal) and tested for each if it can be connected to the roadmap.
Table~\ref{tbl:connectivity} reports on the probability to connect such a random query point to each roadmap. 
As SPANNER does not remove vertices, the probability to connect a random point to the new roadmap does not change. Hence, we only present comparison results of RSEC and the original roadmap.
One can see that for all test cases, there is a negligible degradation in connectivity (if any).

\begin{table}[t,b]
\begin{center}
	\begin{tabular}{|c|c|c|c|c|}
		\hline
							& \multicolumn{2}{c|}{Normal setting } 
							& \multicolumn{2}{c|}{Dense setting }\\ 
		\cline{2-5}
		Scenario 			&	PRM* 		&	RSEC 		&	PRM* 		&	RSEC \\
		\hline
		Easy 					& 100\% 	& 100\% 	& 100\%		& 99.9\% \\
		Cubicles 			& 99.9\% 	& 97.2\% 	& 100\%		& 97.6\% \\
		Bug trap 			& 100\% 	& 100\% 	& 100\%		& 100\% \\
		Alpha Puzzle	& 100\% 	& 100\% 	& 100\%		& 99.9\% \\
		\hline
	\end{tabular}
\end{center}
\caption{\sf  Probability to connect a random point to the roadmap. 
							The drift bound used by RSEC is $d = 0.16$.}
\vspace {-9mm}
\label{tbl:connectivity}
\end{table}

\begin{figure}[b]
	\vspace{-6mm}
  \centering
  \subfloat
   [\sf RSEC, normal setting ]
   {
    \includegraphics[width=0.22\textwidth]{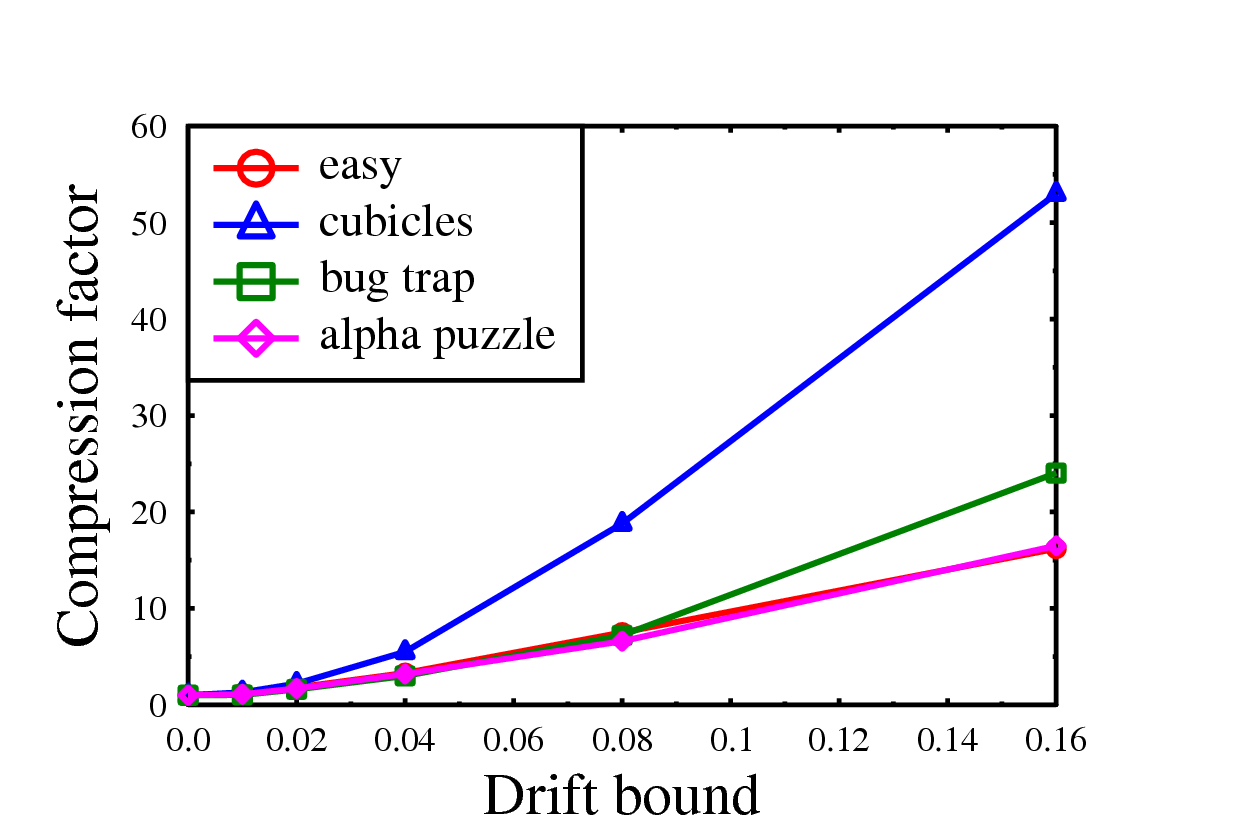}
   }
   \hspace{1mm}
  \subfloat
  [\sf RSEC, dense setting]
   {
    \includegraphics[width=0.22\textwidth]{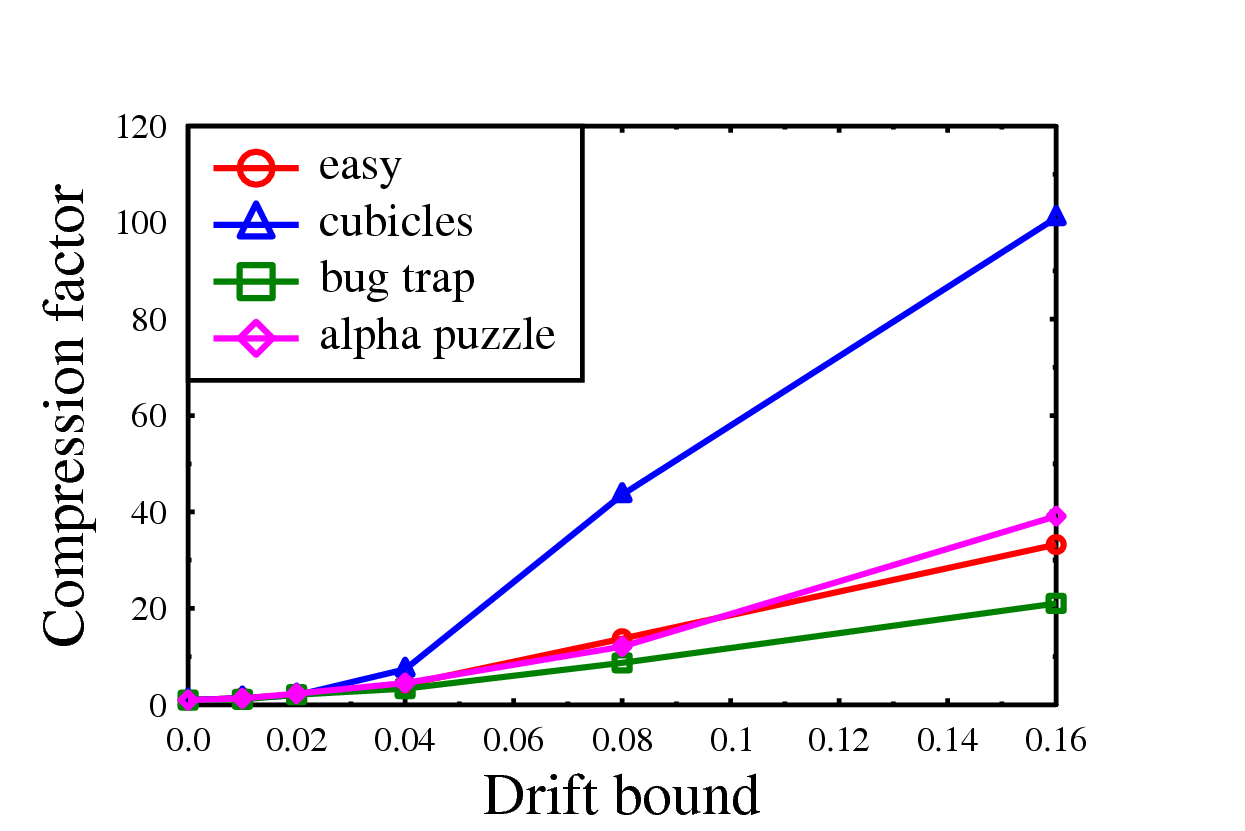}
   }
   \hspace{1mm}
  \subfloat
   [\sf SPANNER, normal setting ]
   {
    \includegraphics[width=0.22\textwidth]{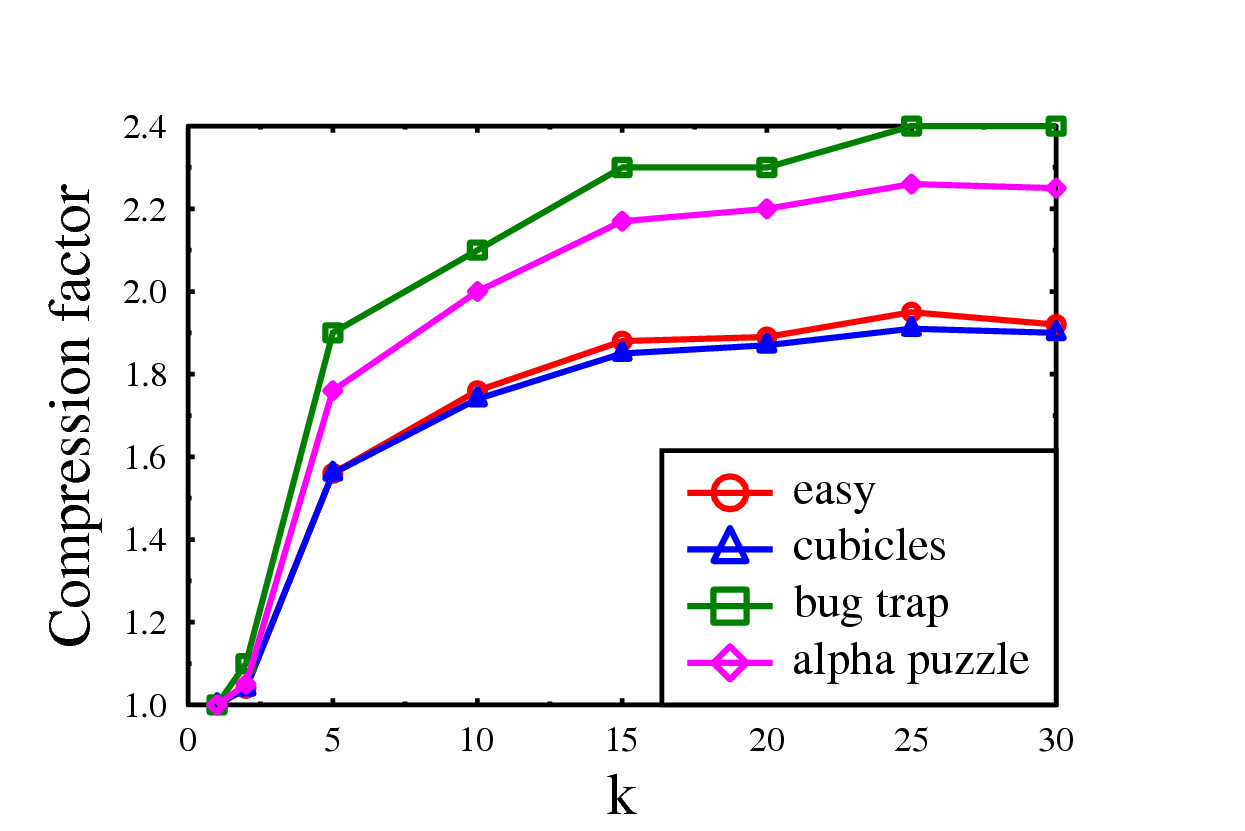}
   }
   \hspace{1mm}
   \subfloat
   [\sf SPANNER, dense setting ]
   {
    \includegraphics[width=0.22\textwidth]{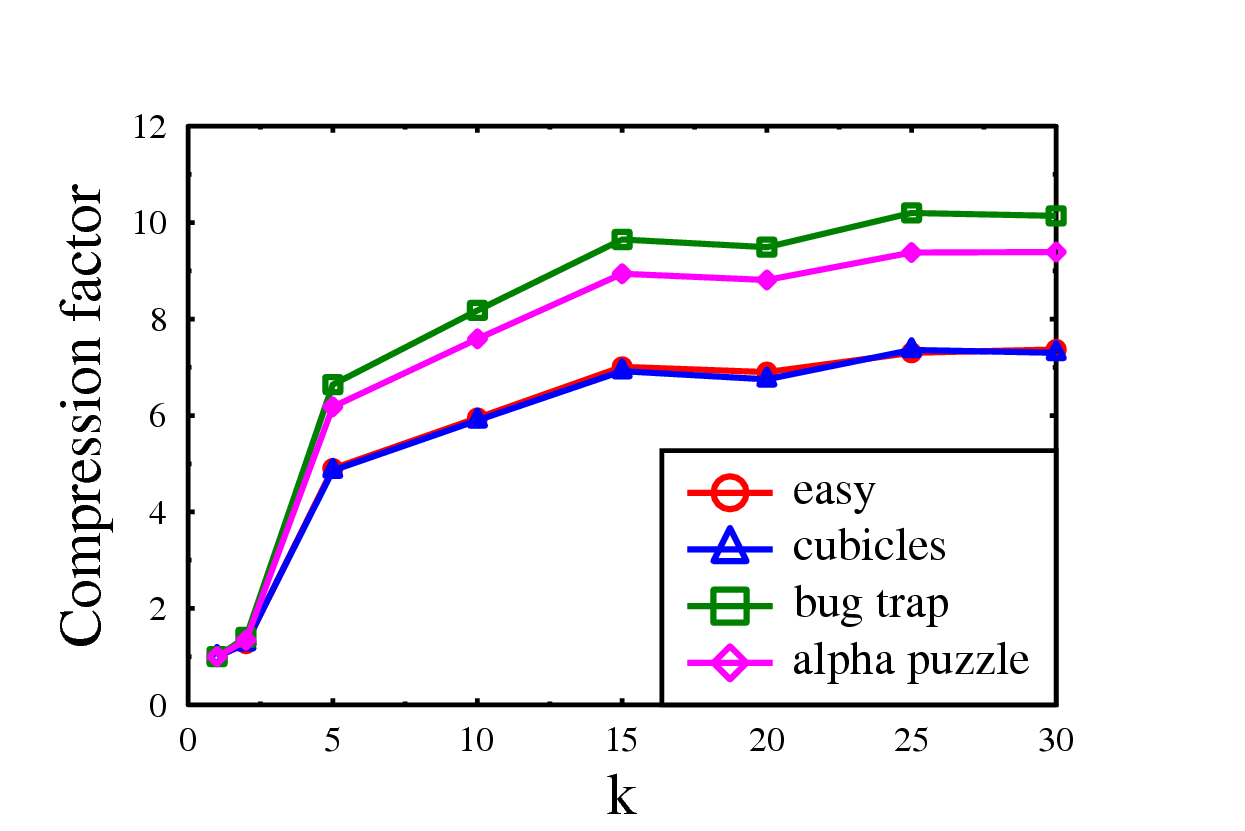}
   }
  \caption{\sf 	Compression factor of the two sparsification algorithms for 
  							each scenario and each setting. Results are averaged over 5 
  							runs.}
  \label{fig:calibration}
\end{figure}

\vspace{3mm}
\textbf{Roadmap Compression}
The amount of compression achieved by each algorithm is governed by its input parameters---$k$ for SPANNER, where $k$ is a parameter related to the stretch, and drift bound for RSEC.
Thus, for each algorithm we plot the compression factor as a function of its input parameter as shown in Figure~\ref{fig:calibration}. 
One can see that the compression factors achieved by RSEC are much higher than SPANNER. Although we ran SPANNER with high values of its input parameter $k$, we did not manage to obtain a higher compression factor than 2.4 for the normal setting and 10 for the dense setting. 
RSEC on the other hand exhibits very high compression factors, up to 55 for the cubicles scenario in the normal setting (this is a sparsified graph which is less than 2\% the size of the original roadmap).

\begin{figure}[t,b]
	\vspace{-10mm}
  \centering
  \subfloat
   [\sf Easy \vspace{-4mm}]
   {
    \includegraphics[width=0.23\textwidth]{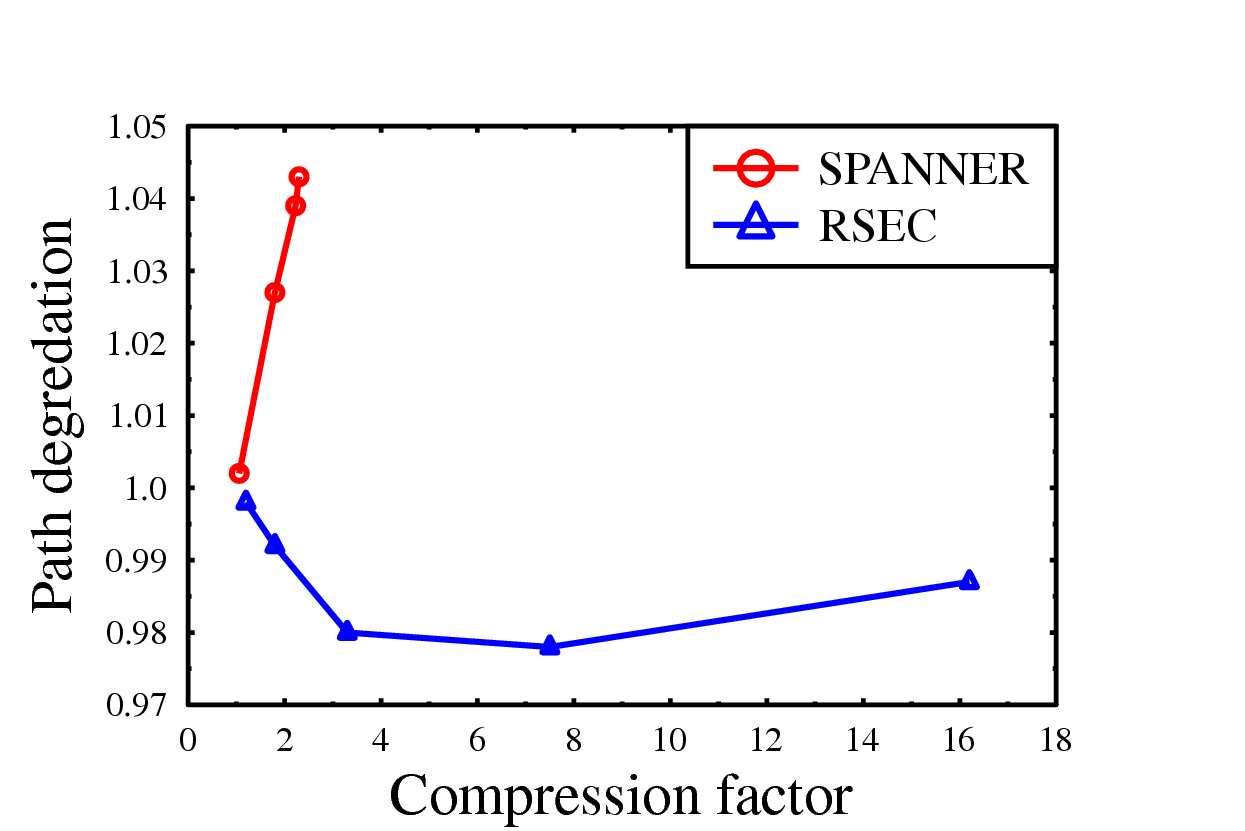}
   }
  \subfloat
   [\sf Cubicles \vspace{-4mm}]
   {
    \includegraphics[width=0.23\textwidth]
    	{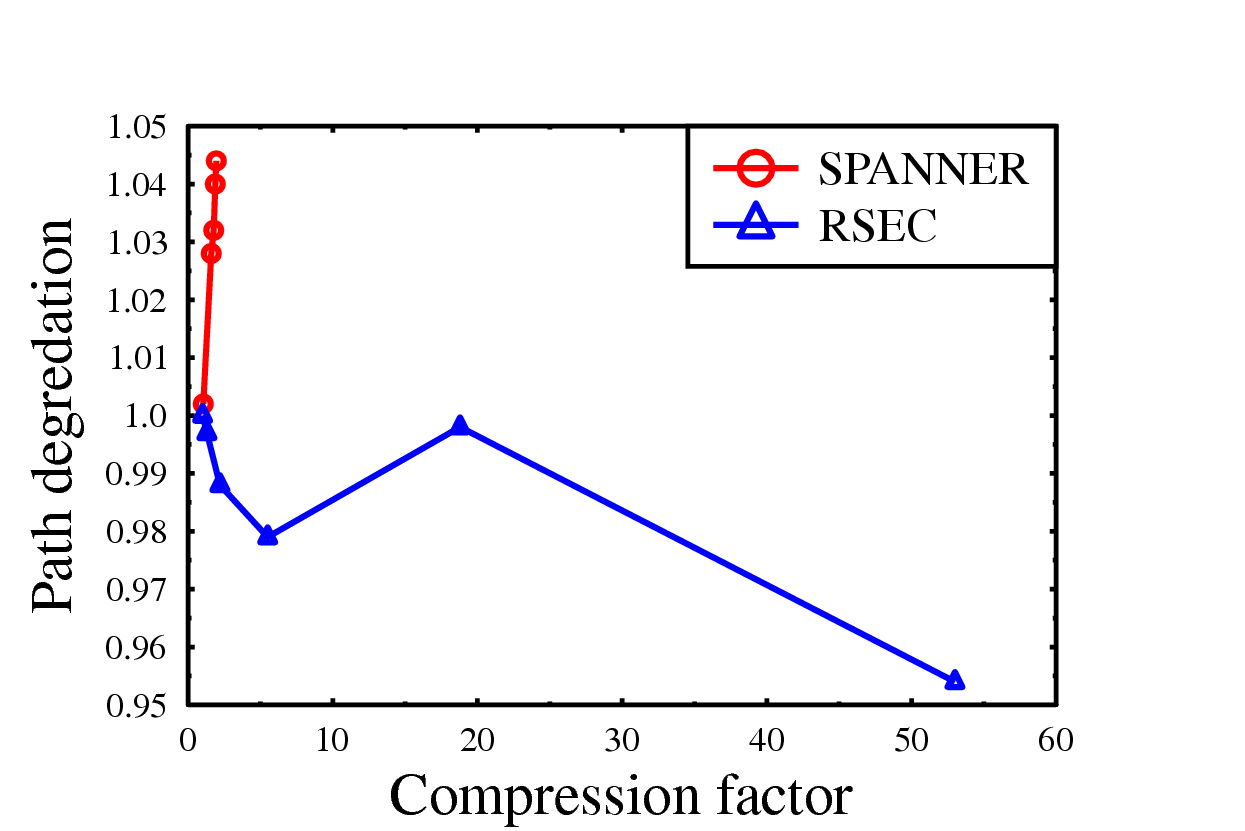}
   }
  \hspace{1mm}
  \subfloat
   [\sf Bug trap]
   {
    \includegraphics[width=0.23\textwidth]{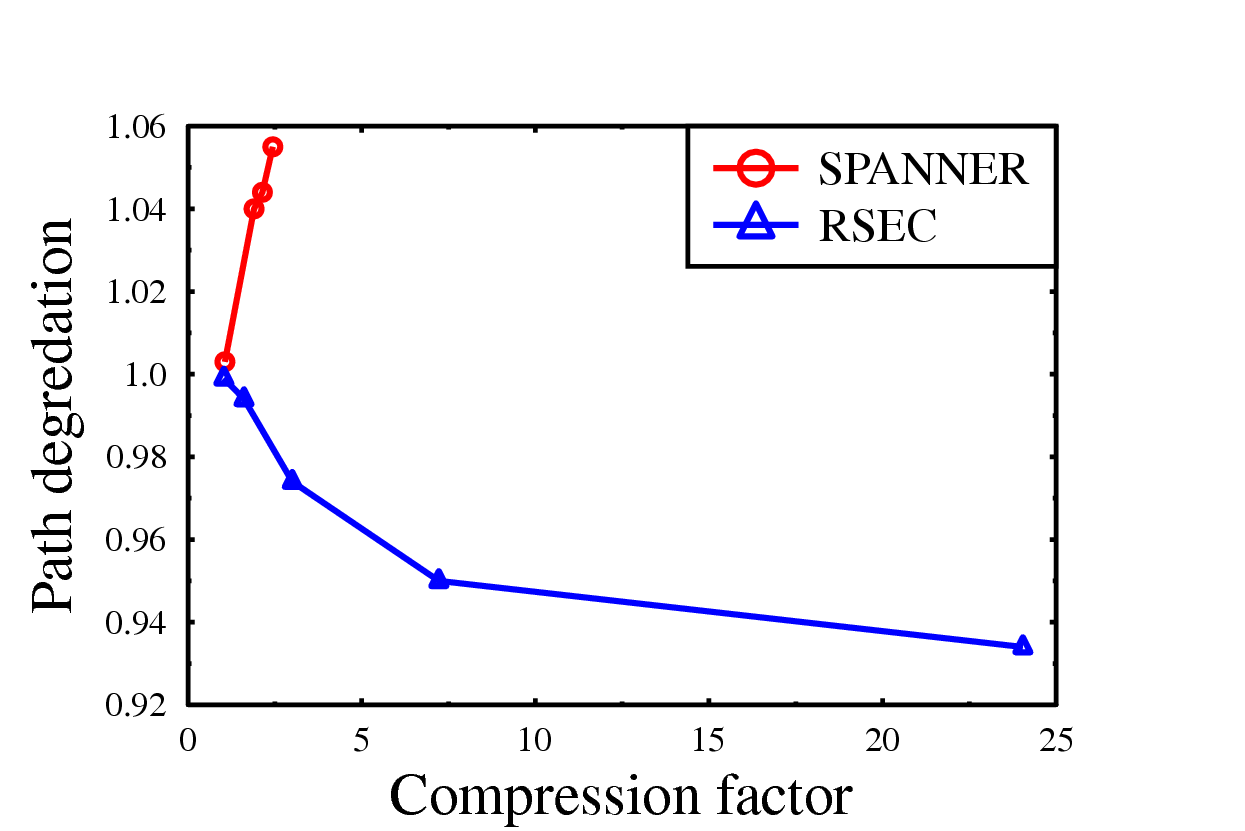}
   }
  \subfloat
   [\sf Alpha puzzle]
   {
    \includegraphics[width=0.23\textwidth]{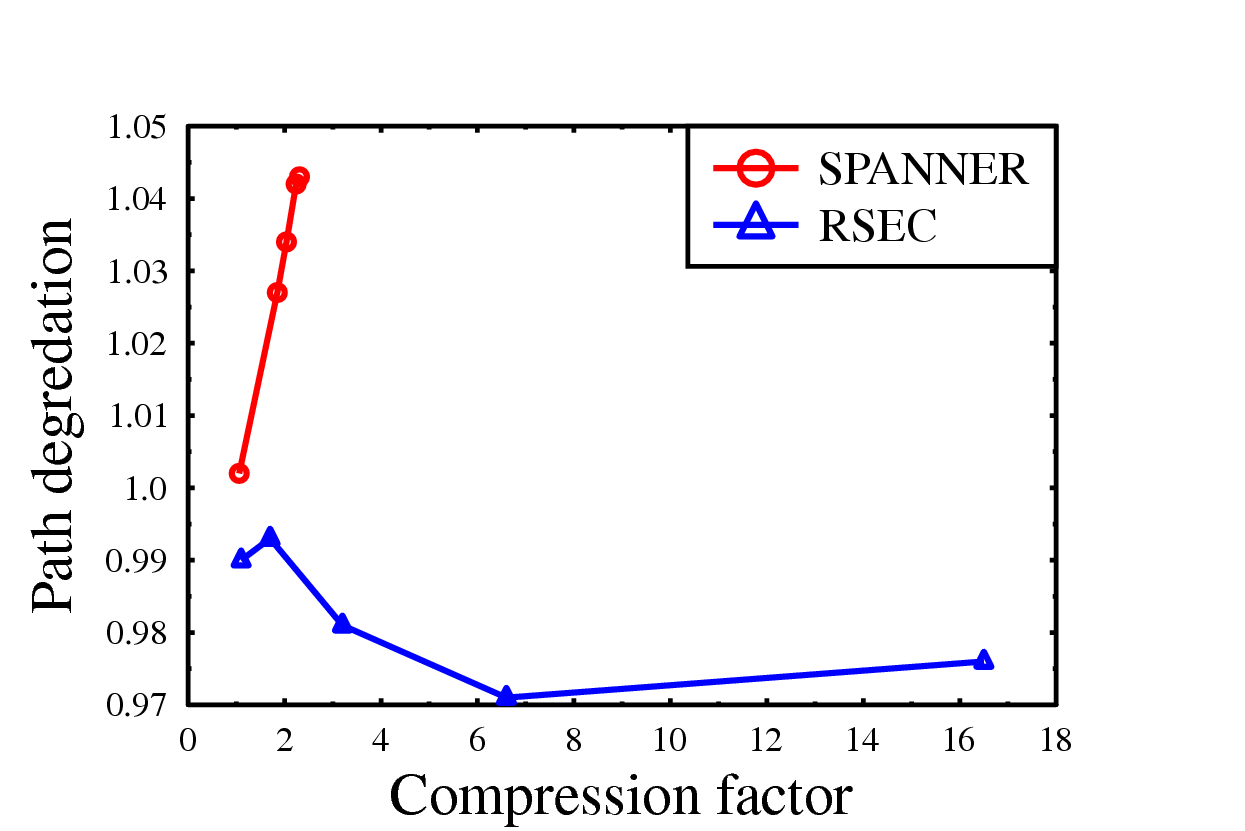}
   }
  \caption{\sf 	Path degradation for each scenario as a function of the 
  							compression factor for each algorithm -- Normal setting. }
  \vspace {-3mm}
  \label{fig:path_degradation_normal}
\end{figure}

\begin{figure}[h,t,b]
	\vspace{-10mm}
  \centering
  \subfloat
   [\sf Easy \vspace{-4mm}]
   {
    \includegraphics[width=0.23\textwidth]{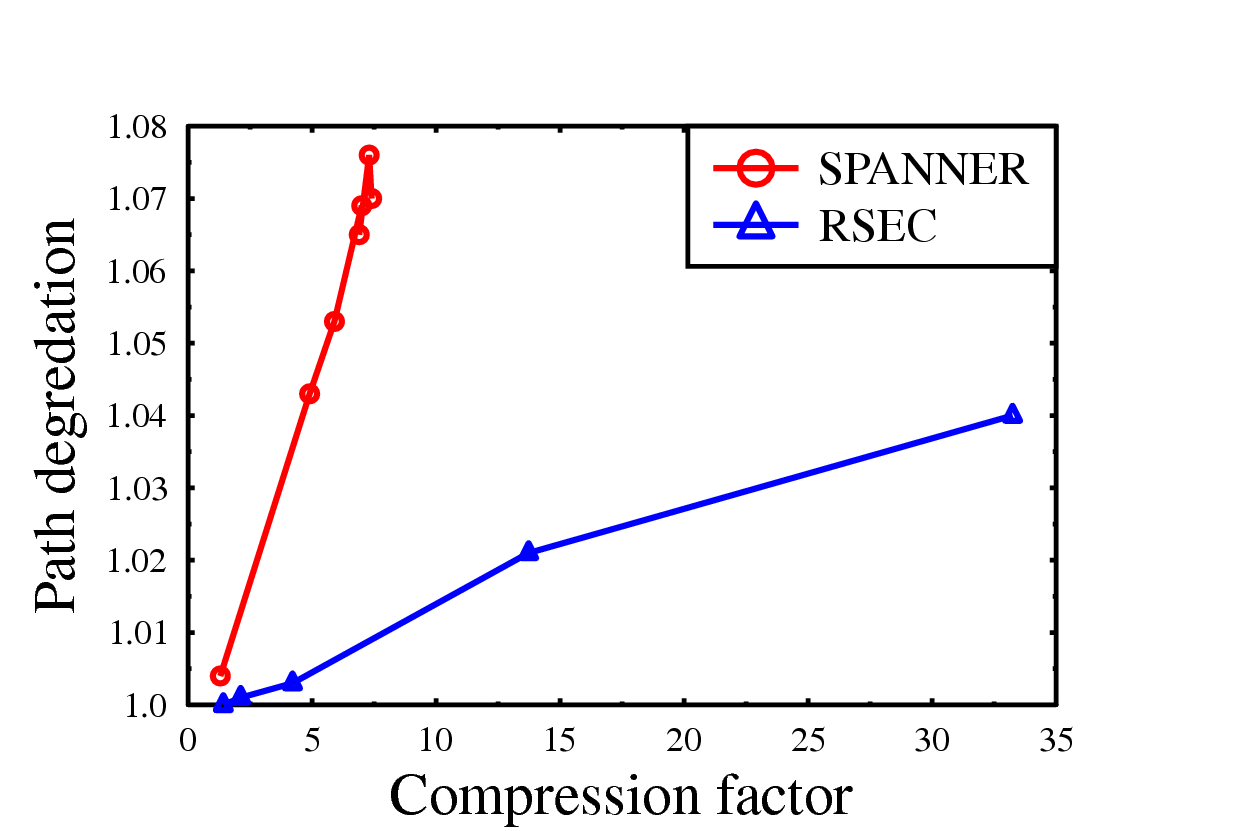}
   }
  \subfloat
   [\sf Cubicles \vspace{-4mm}]
   {
    \includegraphics[width=0.23\textwidth]{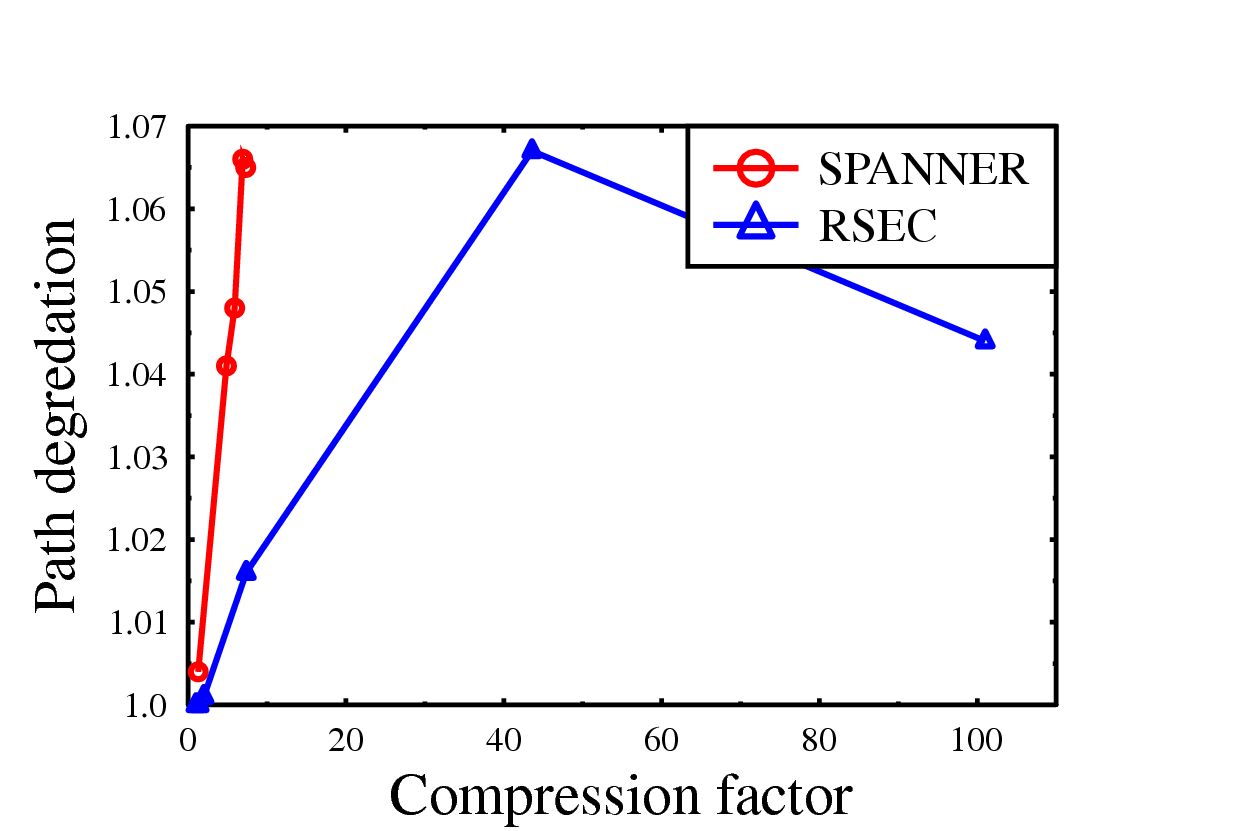}
   }
  \hspace{1mm}
  \subfloat
   [\sf Bug trap]
   {
    \includegraphics[width=0.23\textwidth]{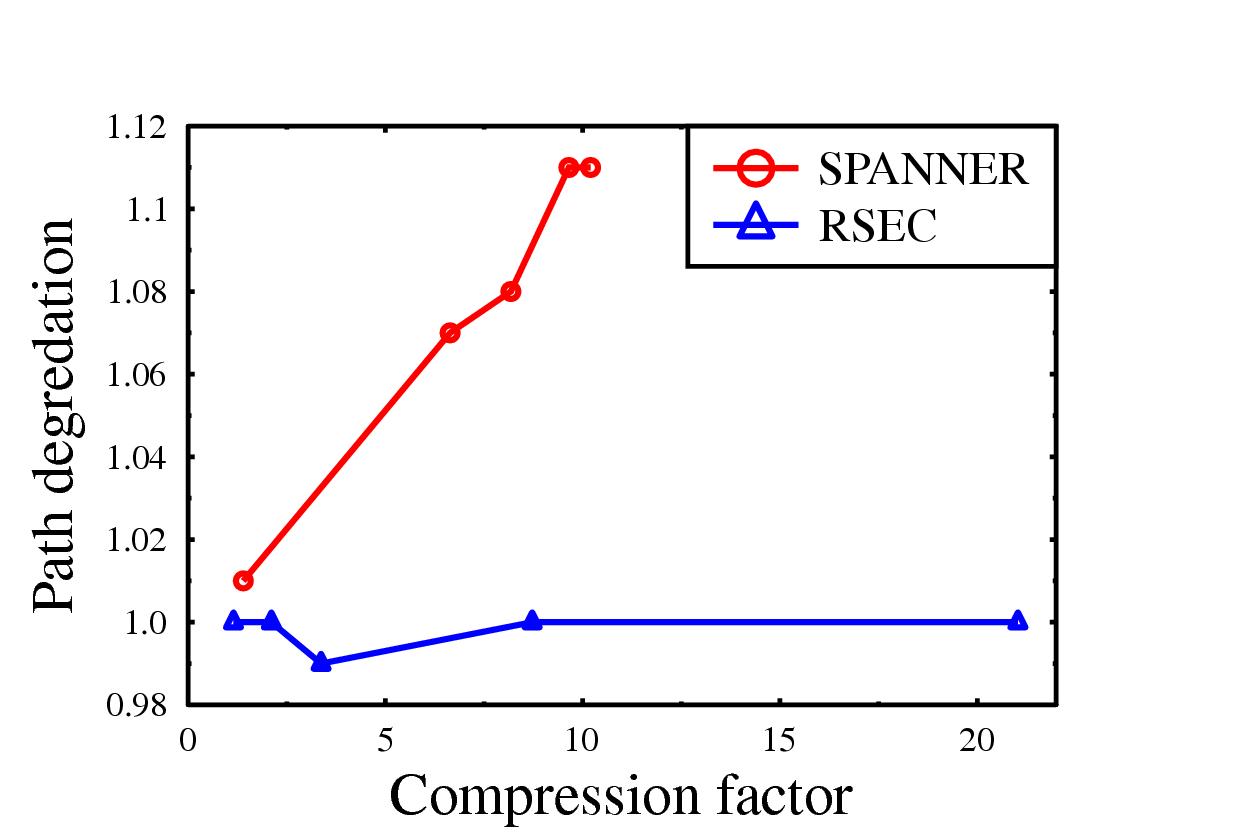}
   }
  \subfloat
   [\sf Alpha puzzle]
   {
    \includegraphics[width=0.23\textwidth]{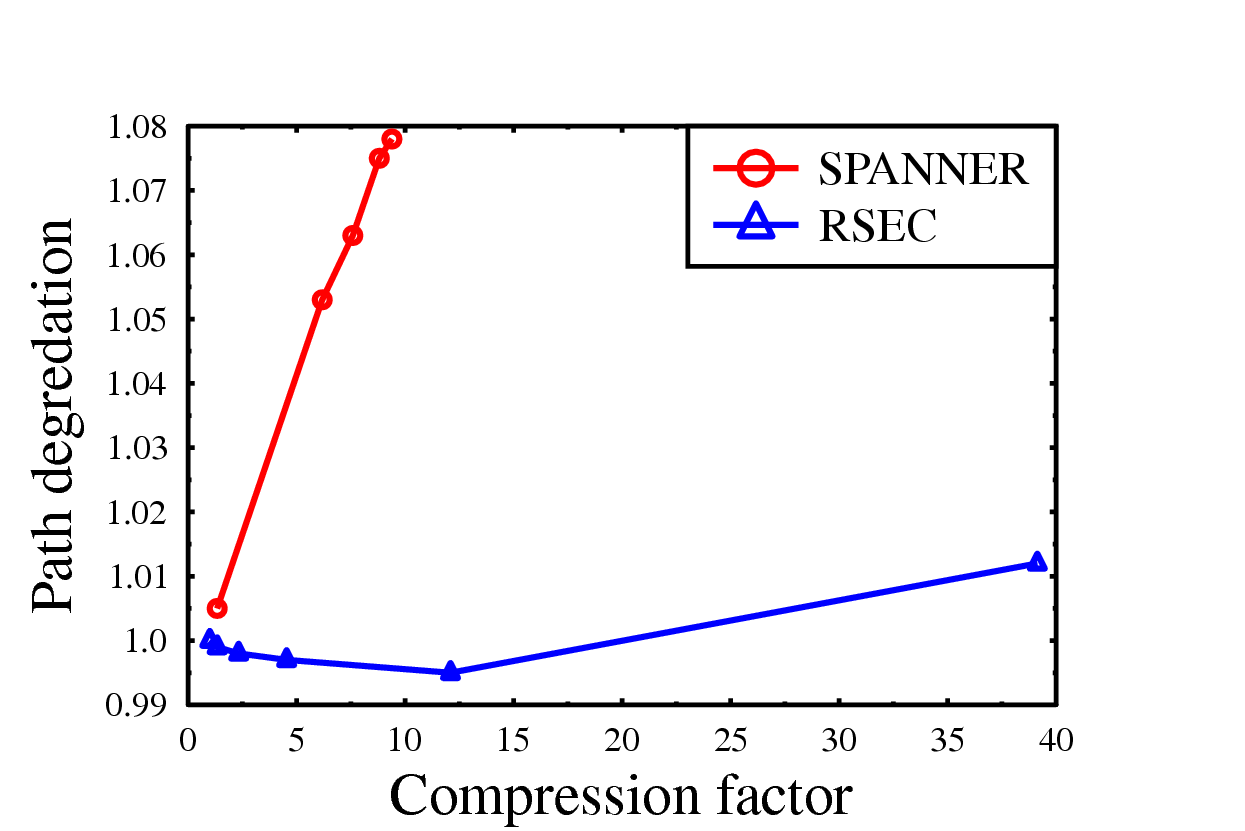}
   }
  \caption{\sf 	Path degradation for each scenario as a function of the 
  							compression factor for each algorithm --- Dense setting. }
  \label{fig:path_degredation_dense}
\end{figure}

\vspace{3mm}
\textbf{Roadmap Quality}
As the compression factor of the two algorithms is based on different parameters, we chose to compare the path quality as a function of the compression factor that was obtained.
Figure~\ref{fig:path_degradation_normal} plots the average degradation in path length as a function of the compression factor for each algorithm for the normal setting. 
For the same values of the compression factor, RSEC exhibits less degradation in average path quality. For example in the bugtrap scenario, for the compression factor of 2.4, SPANNER exhibits a degradation in average path quality of around 5\% while RSEC slightly improves the average quality of paths.
The same behavior is observed for the dense setting depicted in Figure~\ref{fig:path_degredation_dense}. 

%% file: discussion.tex
\section{Discussion}
\label{sec:discussion}
Comparing the difference between RSEC and SPANNER helps understand both the advantages as well as the shortcomings of RSEC.
A key drawback of SPANNER is that it overlooks the fact that the graph on which it operates is a roadmap. Thus, it does not introduce any new vertices and edges and is only capable of removing edges while leaving the set of vertices intact.
In contrast, RSEC makes use of the fact that the roadmap is an approximation of some space and uses both a \emph{collision detector} and a \emph{local planner} to perform edge contraction. 
Clearly these operations are much more time-consuming than edge removal operations, but in return these natural operations in motion-planning algorithms grant RSEC much more power as demonstrated in Section~\ref{sec:experimental results}.

The additional running time incurred by the geometric operations seems reasonable as RSEC is expected to run in an offline phase where running time is not the primary concern.

We saw that the performance of RSEC clearly surpasses SPANNER for graphs with a small average degree. This seems natural as the notion of edge contraction is borrowed from mesh simplifications where the average degree is at most 6. 
As the roadmaps become dense the advantage of RSEC diminishes. 
A possible approach may be to run the SPANNER algorithm with a small stretch factor in order to reduce the average degree (this is bound to happen as the number of vertices does not change) and then run RSEC. 

Additional suggestions for further research include 
(i)~deriving theoretical bounds to the quality of the obtained roadmap and
(ii)~relaxing the edge contraction operation to allow edge contractions even when not all neighbors can be connected to the contraction point. This may lead to higher compressions with limited effect on the path degradation.

%% file: additional_experimntal_results.tex
We include additional experimental results regarding the different heuristics evaluated.
Tables~\ref{tbl:heuristic_comprisons_FIFO},~\ref{tbl:heuristic_comprisons_compresability} and~\ref{tbl:heuristic_comprisons_clearance}  
summarize RSEC's results on all scenarios for the FIFO, Compresability and Clearance heuristic, respectively.

\begin{table*}[t,b]
\footnotesize
\begin{center}
	\begin{tabular}{|c c||c c c c c c||}
		\hline
		&	& &\multicolumn{5}{c||}{drift bound}\\
		Scenario 			
		&	&	original &	$d=0.1$ 	&	$d=0.2$	& $d=0.4$	& $d=0.8$ & $d=0.16$\\
		\hline
		{\multirow{4}{*}{Easy}} 
		&	average path length&  314.8&
								99.9\%	&	99.5\%		&	98.7\%	&	98.8\%	&	99.4\%	\\
		&	compression factor&		1&
								1.16		&	1.8		 		&	3.5			&	9.11 		&	23.6		\\
		&	vertices					&		5016&
								87.5\%	&	61.9\%		&	30.5\%	&	11.4\% 	&	5.1\%		\\
		&	edges							&		48866&
								85.5\%	&	54.3\%		&	28.2\%	&	10.9\% 	&	4.1\%		\\
		\hline
		{\multirow{4}{*}{Cubicles}} 
		&	average path length&  472.4&
								99.9\%	&	99.7\%	&	99.5\%	&	101.8\%	&	103.6\%	\\
		&	compression factor&	  1&
								1.32		&	2.2 		&	5.62 		&	20.42		&	52.23		\\
		&	vertices					&	  5004&
								78.3\%	&	50.1\%	&	20.0\%	&	6.7\% 	&	3.2\%		\\
		&	edges							&	  45217&
								75.4\%	&	44.3\%	&	17.3\%	&	4.5\%		&	1.6\%		\\
		\hline
		{\multirow{4}{*}{Bug trap}} 
		&	average path length&	  41.8&
								100\%		&	99.6\%	&	98.1\%	&	95.8\%	&	95.4\%	\\
		&	compression factor&	  1&
								1.03		&	1.58 		&	3.18		&	9.03		&	29.5		\\
		&	vertices					&	  5011&
								95.7\%	&	69.5\%	&	35.2\%	&	13.0\% 	&	4.8\%		\\
		&	edges							&  66053&
								97.0\%	&	62.6\%	&	30.9\%	&	10.8\%	&	3.2\%		\\		\hline
		{\multirow{4}{*}{Alpha puzzle}} 
		&	average path length&  127.8&
								99.9\%	&	99.5\%	&	98.6\%	&	98.0\%	&	98.4\%  \\
		&	compression factor&  1&
								1.09		&	1.73 		&	3.47 		&	8.51 		&	23.76	\\
		&	vertices					&  5053&
								92.2\%	&	63.7\%	&	30.6\%	&	11.8\%	&	4.5\%		\\
		&	edges							&	  62294&
								91.6\%	&	56.8\%	&	28.5\%	&	17.7\%	&	4.2\%		\\
		\hline		
	\end{tabular}		
\end{center}
\caption{\sf  Average path degradation and compression factor of RSEC 	
							with respect to the original graph using the FIFO 
							heuristic.
							The results are averaged over 5 runs and 150 random queries.}
\label{tbl:heuristic_comprisons_FIFO}
\end{table*}


\begin{table*}[t,b]
\footnotesize
\begin{center}
	\begin{tabular}{|c c||c c c c c c||}
		\hline
		&	& &\multicolumn{5}{c||}{drift bound}\\
		Scenario 			
		&	&	original &	$d=0.1$ 	&	$d=0.2$	& $d=0.4$	& $d=0.8$ & $d=0.16$\\
		\hline
		{\multirow{4}{*}{Easy}} 
		&	average path length&  314.8&
								99.9\%	&	99.4\%		&	98.8\%	& 99.7\%	&101.1\%	\\
		&	compression factor&		1&
								1.18		&	1.82	 		&	3.43		&	8.01		&	16.0		\\
		&	vertices					&		5016&
								8703\%	&	62.1\%		&	32.7\%	&	15.0\% 	&	9.1\%		\\
		&	edges							&		48866&
								84.4\%	&	53.5\%		&	28.4\%	&	12\% 		&	5.7\%		\\
		\hline
		{\multirow{4}{*}{Cubicles}} 
		&	average path length&  472.4&
								99.9\%	&	99.4\%	& 99.9\%	&	103.0\%	&	104.5\%	\\
		&	compression factor&	  1&
								1.34		&	2.21		&	5.21 		&	13.34		&	29.14		\\
		&	vertices					&	  5004&
								77.7\%	&	51.4\%	&	23.6\%	&	10.8\% 	&	6.0\%		\\
		&	edges							&	  45217&
								74.0\%	&	44.0\%	&	18.2\%	&	6.8\%		&	2.9\%		\\
		\hline
		{\multirow{4}{*}{Bug trap}} 
		&	average path length&	  41.8&
								99.9\%		&	99.7\%	&	97.7\%	&	95.9\%	&	94.8\%	\\
		&	compression factor&	  1&
								1.03		&	1.63 		&	3.18	  &	8.42		&	17.82		\\
		&	vertices					&	  5011&
								95.7\%	&	67.5\%	&	34.8\%	&	14.2\% 	&	8.2\%		\\
		&	edges							&  66053&
								97.0\%	&	60.2\%	&	30.9\%	&	11.5\%	&	5.2\%		\\		\hline
		{\multirow{4}{*}{Alpha puzzle}} 
		&	average path length&  128.2&
								99.9\%	&	99.6\%	&	99.5\%	&	99.4\%	&	101.6\%  	\\
		&	compression factor&  1&
								1.09		&	1.82		&	4.28		&	12.25		&	31.29			\\
		&	vertices					&  5053&
								92.3\%	&	64.1\%	&	31.3\%	&	12.3\%	&	6.3\%			\\
		&	edges							&	  62294&
								91.3\%	&	53.6\%	&	22.0\%	&	7.5\%		&	2.7\%			\\
		\hline		
	\end{tabular}		
\end{center}
\caption{\sf  Average path degradation and compression factor of RSEC 	
							with respect to the original graph using the Compressibility  
							heuristic.
							The results are averaged over 5 runs and 150 random queries.}
\label{tbl:heuristic_comprisons_compresability}
\end{table*}


\begin{table*}[t,b]
\footnotesize
\begin{center}
	\begin{tabular}{|c c||c c c c c c||}
		\hline
		&	& &\multicolumn{5}{c||}{drift bound}\\
		Scenario 			
		&	&	original &	$d=0.1$ 	&	$d=0.2$	& $d=0.4$	& $d=0.8$ & $d=0.16$\\
		\hline
		{\multirow{4}{*}{Easy}} 
		&	average path length&  314.8&
								99.9\%	&	99.7\%		&	99.7\%	&101.4\%	&104.4\%	\\
		&	compression factor&		1&
								1.18		&	1.89	 		&	4.25		&	11.44		&	26.27		\\
		&	vertices					&		5016&
								87.3\%	&	63.1\%		&	32.0\%	&	13.8\% 	&	8.1\%		\\
		&	edges							&		48866&
								84.3\%	&	50.8\%		&	21.8\%	&	7.7\% 	&	2.9\%		\\
		\hline
		{\multirow{4}{*}{Cubicles}} 
		&	average path length&  472.4&
								100\%	&	99.7\%	&100.4\%	&	104.0\%	&	106.0\%	\\
		&	compression factor&	  1&
								1.36		&	2.43		&	6.79 		&	21.73		&	45.0		\\
		&	vertices					&	  5004&
								77.7\%	&	50.7\%	&	22.3\%	&	8.9\% 	&	5.6\%		\\
		&	edges							&	  45217&
								72.7\%	&	39.0\%	&	13.0\%	&	3.7\%		&	1.5\%		\\
		\hline
		{\multirow{4}{*}{Bug trap}} 
		&	average path length&	  41.8&
								100\%		&	99.8\%	&	98.7\%	&	96.9\%	&	97.5\%	\\
		&	compression factor&	  1&
								1.03		&	1.65 		&	3.94		&	12.71		&	41.27		\\
		&	vertices					&	  5011&
								95.8\%	&	69.1\%	&	34.9\%	&	12.8\% 	&	5.3\%		\\
		&	edges							&  66053&
								96.9\%	&	59.1\%	&	24.0\%	&	7.11\%	&	2.0\%		\\		\hline
		{\multirow{4}{*}{Alpha puzzle}} 
		&	average path length&  127.8&
								99.7\%	&	99.2\%	&	98.3\%	&	97.4\%	& 97.4\%  \\
		&	compression factor&  1&
								1.1 		&	1.8 		&	3.73 		&	9.43		&	24.18	\\
		&	vertices					&  5053&
								91.7\%	&	61.7\%	&	29.0\%	&	11.1\%	&	5.3\%		\\
		&	edges							&	  62294&
								91.1\%	&	54.8\%	&	26.4\%	&	10.5\%	&	4.0\%		\\
		\hline		
	\end{tabular}		
\end{center}
\caption{\sf  Average path degradation and compression factor of RSEC 	
							with respect to the original graph using the Clearance 
							heuristic.
							The results are averaged over 5 runs and 150 random queries.}
\label{tbl:heuristic_comprisons_clearance}
\end{table*}